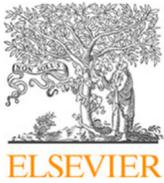
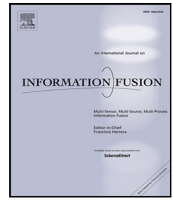

Full length article

# Error adjustment based on spatiotemporal correlation fusion for traffic forecasting

Fuqiang Liu [a],[*], Weiping Ding [b], Luis Miranda-Moreno [a], Lijun Sun [a],[*]

[a] *Department of Civil Engineering, McGill University, Montreal, H3A 0G4, Quebec, Canada*
[b] *School of Articial Intelligence and Computer Science, Nantong University, Nantong, 226019, China*

ABSTRACT

Deep neural networks (DNNs) play a significant role in an increasing body of research on traffic forecasting due to their effectively capturing spatiotemporal patterns embedded in traffic data. A general assumption of training the said forecasting models via mean squared error estimation is that the errors across time steps and spatial positions are uncorrelated. However, this assumption does not really hold because of the autocorrelation caused by both the temporality and spatiality of traffic data. This gap limits the performance of DNN-based forecasting models and is overlooked by current studies. To fill up this gap, this paper proposes Spatiotemporally Autocorrelated Error Adjustment (SAEA), a novel and general framework designed to systematically adjust autocorrelated prediction errors in traffic forecasting. Unlike existing approaches that assume prediction errors follow a random Gaussian noise distribution, SAEA models these errors as a spatiotemporal vector autoregressive (VAR) process to capture their intrinsic dependencies. First, it explicitly captures both spatial and temporal error correlations by a coefficient matrix, which is then embedded into a newly formulated cost function. Second, a structurally sparse regularization is introduced to incorporate prior spatial information, ensuring that the learned coefficient matrix aligns with the inherent road network structure. Finally, an inference process with test-time error adjustment is designed to dynamically refine predictions, mitigating the impact of autocorrelated errors in real-time forecasting. The effectiveness of the proposed approach is verified on different traffic datasets. Results across a wide range of traffic forecasting models show that our method enhances performance in almost all cases.

## 1. Introduction

The fusion of spatial and temporal information is essential for accurate traffic forecasting, a long-standing research challenge and a fundamental component of intelligent transportation systems (ITS) [1–3]. In recent years, deep neural network (DNN)-based forecasting models, particularly those utilizing graph neural networks (GNNs), have made significant advancements in traffic forecasting. Notable examples include DCRNN [4], STGCN [5], Graph WaveNet [6], and Gman [7]. These models effectively capture complex spatiotemporal dependencies by leveraging both traffic flow dynamics and road network structures, leading to enhanced forecasting accuracy. As a result, they have demonstrated superior predictive performance, making them highly effective for real-world traffic forecasting applications [8].

During the training of DNN-based models, the mean square error (MSE) or mean average error (MAE) loss function is utilized [9–11], implicitly assuming that prediction errors exhibit no autocorrelation and behave as independent and identically distributed (i.i.d.) white noise. However, recent studies [12] have demonstrated that this assumption can substantially degrade the performance of deep learning-based univariate time series models. Moreover, the findings suggest that explicitly accounting for error autocorrelation in the modeling process can improve forecasting accuracy, highlighting the importance of reconsidering traditional loss functions in spatiotemporal learning.

We argue that the assumption of no autocorrelation in errors, which underlies the use of the MSE/MAE loss function in training traffic forecasting models, can significantly degrade their predictive performance. Traffic state variables (e.g., speed and flow) are typically represented as multivariate time series collected from fixed sensors (e.g., loop detectors) deployed throughout road networks [13–15]. Autocorrelated errors are particularly prevalent in traffic data due to spatiotemporally varying factors, such as weather conditions, traffic control interventions (e.g., signal timing adjustments), traffic demand fluctuations, and congestion propagation. These factors are often treated as external variables and are not directly incorporated into DNN-based forecasting models. Furthermore, model misspecification [16,17] can exacerbate error autocorrelation by failing to adequately capture the underlying data-generating process (DGP) governing traffic dynamics. In summary, autocorrelated errors are more pronounced in spatiotemporal traffic data compared to univariate time series, making the no-autocorrelation assumption more detrimental to the performance of DNN-based traffic forecasting models.

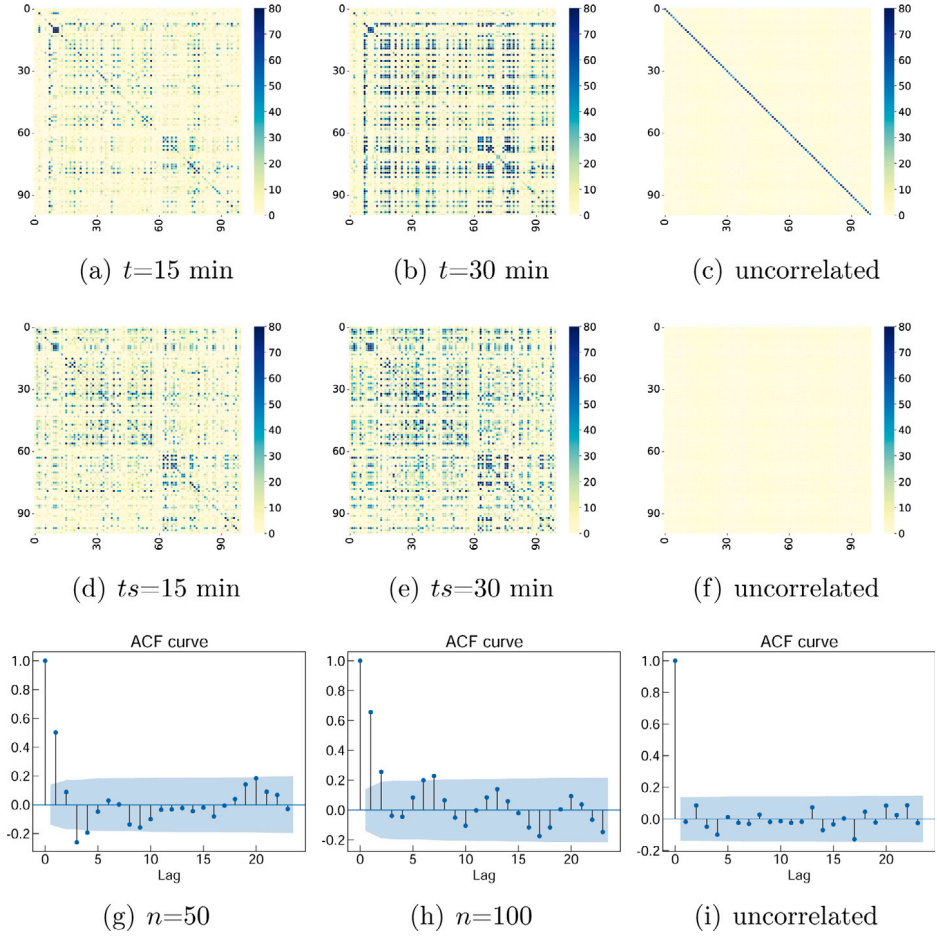

**Fig. 1.** Spatiotemporal correlation in traffic forecasting errors. Supposing $\epsilon_t$ is an $N \times 1$ vector of prediction errors, panels (a–b) present spatial residual covariance matrices, $E(\epsilon_t \epsilon_t^T)$, for three/six-step-ahead forecasts ($t = 15/30$ min). Ideally, if prediction errors were spatially independent, these matrices would be diagonal, as in panel (c). Panels (d–e) display the cross-lag covariance matrices, $E(\epsilon_t \epsilon_{t+ts}^T)$, for time lags of $ts = 15/30$ min in three-step-ahead forecasting $t = 15$ min. If prediction errors were truly spatiotemporally uncorrelated, all entries in these matrices would be close to zero, similar to panel (f). Panels (g–h) present the autocorrelation function (ACF) plots for sensors #50 and #100, respectively. If prediction errors were temporally uncorrelated, all spikes, except the one at lag 0, would remain within the threshold shadow, as in panel (i).

Fig. 1 illustrates the empirical correlation structure of forecasting errors when applying STGCN [5] to the PeMS dataset. Based on the forecasting results, we empirically quantify $E(\epsilon_t \epsilon_t^T)$ for spatial residual covariances, $E(\epsilon_t \epsilon_{t+ts}^T)$ for cross-lag covariances, and the autocorrelation function (ACF) curves of two specific sensors, where $\epsilon_t$ represents an $N \times 1$ vector of prediction errors, and $N$ denotes the number of sensors. In the absence of spatial or temporal correlation in forecasting errors, the spatial residual covariance matrix should be diagonal, as shown in Fig. 1(c), the cross-lag covariance matrix should approximate a zero matrix, as depicted in Fig. 1(f), and the ACF curve should resemble Fig. 1(i). However, the left two columns in Fig. 1 empirically reveal a clear spatiotemporal correlation structure in prediction errors, indicating that forecasting errors are neither spatially nor temporally independent. These analyses collectively highlight the presence of spatial and temporal dependencies in forecasting errors, challenging the assumption of independent error distributions in traffic forecasting models. Addressing error autocorrelation in model design and training could potentially enhance the performance of DNN-based traffic forecasting models by incorporating spatiotemporal dependencies into the learning process.

To address the issue of correlated residuals in deep learning-based univariate time series forecasting models, recent studies [12] recently proposed a simple yet effective adjustment for autocorrelated errors in the temporal dimension. This approach introduces a first-order autoregressive error term and jointly leans the autocorrelation coefficient alongside other deep learning model parameters, leading to improved predictive performance across various deep learning-based time series forecasting models. However, while this method effectively mitigates temporally autocorrelated errors in univariate forecasting, its applicability to spatiotemporal forecasting remains unclear. Traffic forecasting involves the fusion of spatial and temporal dependencies, making it essential to develop methodologies that explicitly account for both spatial and temporal error correlations.

In this work, we extend the concept of adjusting for autocorrelated errors to the setting of multivariate and spatiotemporal traffic forecasting, where unexplained error autocorrelation is often more prevalent and pronounced across both temporal and spatial dimensions. Spatiotemporally Autocorrelated Error Adjustment (SAEA) is proposed for DNN-based traffic forecasting. We first employ a vector autoregressive (VAR) process to characterize the spatiotemporal error autocorrelation for multivariate traffic time series. The coefficient matrix is introduced to capture the spatiotemporal correlation of forecasting errors. Second, the coefficient matrix is embedded into the cost function. With the reformed cost function, the forecasting model's parameters along with the coefficient matrix are learned jointly, and spatiotemporally autocorrelated errors can be dismissed by a simple and clear procedure. Finally, structurally sparse regularization is proposed to embed the prior spatial structure into the coefficient matrix and interpret how prediction errors are spatiotemporally correlated. The proposed SAEA only reforms the cost function in the training process, thus it can enhance amounts of traffic forecasting models without any inside modifications.

Through extensive large-scale experiments, we demonstrate that the proposed SAEA framework significantly enhances the forecasting performance of various DNN-based architectures while improving their robustness to spatiotemporal error dependencies. Specifically, SAEA achieves an average prediction accuracy improvement of 6.5% across five state-of-the-art traffic forecasting models, highlighting its effectiveness in mitigating autocorrelated prediction errors. Unlike existing approaches that focus on specific architectures, SAEA is designed to be model-agnostic, and it effectively captures and adjusts prediction errors that exhibit strong autocorrelation across different time steps and locations in the traffic network.

In general, the contributions of this paper can be summarized:

- Error Modeling with a Spatiotemporal VAR Process: Unlike existing methods that assume independent error distributions, we model forecasting errors as a spatiotemporal vector autoregressive (VAR) process, explicitly capturing dependencies across time and space.
- Coefficient Matrix for Error Correlation Modeling: We introduce a coefficient matrix that encodes spatiotemporal correlations in forecasting errors and embed it into a reformed cost function, allowing the model to jointly learn both forecasting parameters and error dependencies.
- Structurally Sparse Regularization for Spatial Embedding: To better integrate prior knowledge of the traffic network, we propose a structurally sparse regularization that constrains the coefficient matrix, improving interpretability and enhancing forecasting robustness.
- Model-Agnostic Error Adjustment Framework: SAEA is designed as a generalizable enhancement for existing deep learning-based traffic forecasting models. It only modifies the cost function without altering model architectures, making it widely applicable to different forecasting frameworks.
- Comprehensive Empirical Validation: We conduct extensive experiments on five state-of-the-art traffic forecasting models across multiple datasets, demonstrating that SAEA improves forecasting accuracy by an average of 6.5%, effectively mitigating autocorrelated errors while preserving computational efficiency.

The rest of this paper is structured as follows: Section 2 introduces DNN-based traffic forecasting, discussing existing methodologies and their limitations. Section 3 details the proposed SAEA framework, including the mathematical formulation and error adjustment methodology. Section 4 presents extensive empirical evaluations, comparing SAEA against baseline approaches and analyzing its effectiveness across different models and datasets. Section 5 concludes the paper with key findings and directions for future research.

## 2. DNN-based traffic forecasting

To better incorporate the spatial road network structure, recent state-of-the-art traffic forecasting models [18–20] primarily rely on the combination of temporal deep networks with graph neural networks [21–23]. These models effectively capture the underlying spatiotemporal sequence from $N$ sensors by representing traffic systems as a time-varying graph.

To model the evolving traffic dynamics, these models represent the traffic system as:

$$\mathcal{G}_t = \{\mathcal{V}_t, \mathcal{E}, \mathcal{W}\} \tag{1}$$

where:

- $\mathcal{V}_t = \{v_{1,t}, \ldots, v_{N,t}\}$ represents the set of traffic states (e.g., speed, flow, or density) at time $t$.
- $\mathcal{E}$ is the set of edges, encoding spatial connectivity between different locations.
- $\mathcal{W}$ is the weighted adjacency matrix, capturing the strength of spatial relationships between sensors in the road network.

By formulating traffic forecasting as a spatiotemporal graph learning problem, these models learn complex interactions between road segments and adapt dynamically to changing traffic conditions [24–26].

Supposing $[\mathcal{G}_{t-1}, \ldots, \mathcal{G}_{t-H}]$ represents the historical data and $f(\cdot)$ denotes the forecasting model parameterized by $\theta$, the task of multi-step traffic forecasting can be formulated as:

$$\{\mathcal{G}^*_{t+W-1}, \ldots, \mathcal{G}^*_t\} = f(\mathcal{G}_{t-1}, \ldots, \mathcal{G}_{t-H}; \theta), \tag{2}$$

where $[\mathcal{G}^*_{t+W-1}, \ldots, \mathcal{G}^*_t]$ denotes the predicted future traffic state. The model takes historical data from $t-H$ to $t-1$ as input and produces a length-$W$ prediction window from $t$ to $t+W-1$ (typically, $W \leq H$). This approach is commonly referred to as sequence-to-sequence (seq2seq) forecasting.

Although intuitive, seq2seq forecasting models often suffer from training difficulties. An alternative approach to multi-step forecasting is recursively applying a well-trained one-step-ahead forecasting model:

$$\begin{cases} \mathcal{G}^*_t = f(\mathcal{G}_{t-1}, \mathcal{G}_{t-2}, \ldots, \mathcal{G}_{t-H}; \theta), \\ \mathcal{G}^*_{t+1} = f(\mathcal{G}^*_t, \mathcal{G}_{t-1}, \ldots, \mathcal{G}_{t-H+1}; \theta), \\ \vdots \\ \mathcal{G}^*_{t+W-1} = f(\mathcal{G}^*_{t+W-2}, \mathcal{G}^*_{t+W-3}, \ldots, \mathcal{G}_{t-H+W-1}; \theta). \end{cases} \tag{3}$$

In addition to these strategies, [27] propose a simple yet effective method that significantly enhances forecasting performance, particularly for long-term traffic prediction. The key idea is to train multiple models with the same structure but optimized for different steps, formulated as:

$$\begin{cases} \mathcal{G}^*_t = f_0(\mathcal{G}_{t-1}, \mathcal{G}_{t-2}, \ldots, \mathcal{G}_{t-H}; \theta_0), \\ \mathcal{G}^*_{t+1} = f_1(\mathcal{G}_{t-1}, \mathcal{G}_{t-2}, \ldots, \mathcal{G}_{t-H}; \theta_1), \\ \vdots \\ \mathcal{G}^*_{t+W-1} = f_{W-1}(\mathcal{G}_{t-1}, \mathcal{G}_{t-2}, \ldots, \mathcal{G}_{t-H}; \theta_{W-1}), \end{cases} \tag{4}$$

where $f_0, f_1, \ldots, f_{W-1}$ denote forecasting models with identical architectures but distinct parameters, $\theta_0, \theta_1, \ldots, \theta_{W-1}$.

State-of-the-art traffic forecasting models, $f(\cdot)$, are typically constructed using spatiotemporal DNNs, incorporating both spatial and temporal layers. For instance, long short-term memory networks (LSTMs) [28], gated recurrent units (GRUs) [29], or gated convolutional networks (Gated-CNN) [30] are often employed as temporal layers, while GNN-based architectures [31,32] serve as spatial layers, enabling the models to capture complex spatiotemporal dependencies in traffic dynamics effectively.

## 3. Autocorrelation adjustment

Most existing DNN-based traffic forecasting models frame the forecasting task as a regression problem and typically assume that prediction errors are independent and identically distributed (i.i.d.). Under this assumption, minimizing the mean squared error (MSE) is mathematically equivalent to maximizing the likelihood (MLE), which justifies the widespread use of MSE or MAE as training objectives. However, in real-world forecasting scenarios, this i.i.d. assumption often breaks down due to factors such as unobserved external influences, measurement noise, and model misspecification [12], leading to both temporal and spatial correlations in prediction errors.

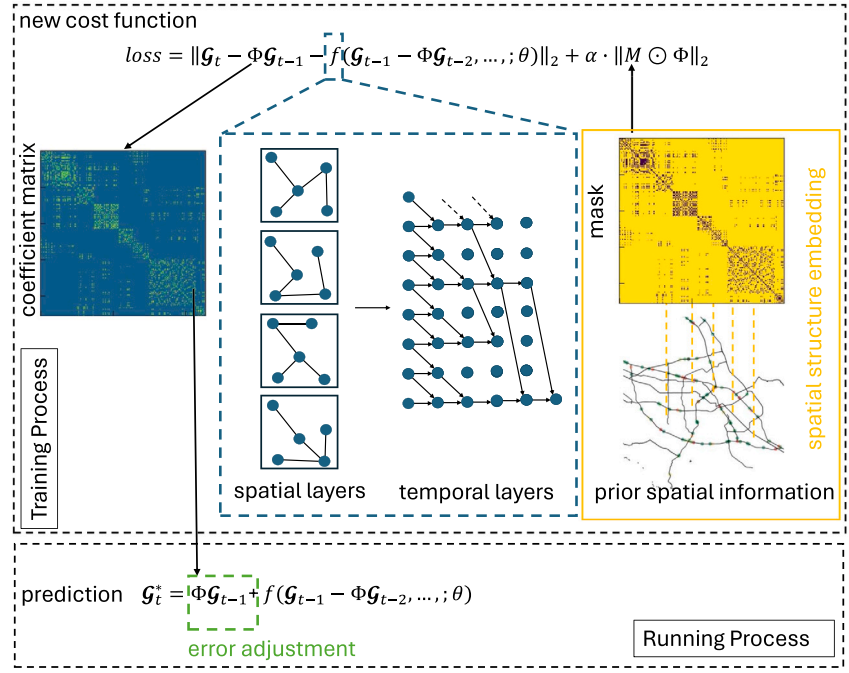

**Fig. 2.** Overview of the proposed SAEA framework. The prediction error is modeled as an autoregressive process, with the coefficient matrix incorporated into the loss function to dismiss the error correction. Additionally, prior spatial knowledge, such as traffic network topology, is embedded in the regularization term to better capture spatial correlations and improve forecasting accuracy.

In contrast to prior studies that rely on the i.i.d. assumption, our work introduces a more comprehensive framework by:

1. Modeling prediction errors as a vector autoregressive (VAR) process rather than an i.i.d. Gaussian noise, and incorporating a coefficient matrix to account for spatiotemporal autocorrelation (Section 3.1).
2. Proposing a novel cost function that adjusts for autocorrelated forecasting errors by embedding the coefficient matrix (Sections 3.2 and 3.3).
3. Introducing spatially sparse regularization to integrate prior structural information into the coefficient matrix, enhancing interpretability and robustness (Section 3.4).

Fig. 2 presents an overview of the proposed SAEA framework, designed to enhance DNN-based traffic forecasting models by mitigating spatiotemporally autocorrelated errors. **The training process**: prior graph structures and spatial information are incorporated to construct a structural mask, which constrains the coefficient matrix. This matrix captures spatiotemporal dependencies in prediction errors and is embedded into a newly designed cost function. Unlike conventional approaches that assume Gaussian white noise, our cost function is formulated based on an autoregressive process, leveraging the learned coefficient matrix to correct for correlated residuals and enhance forecasting accuracy. **The running process**, the learned coefficient matrix is applied to the model's predictions through a forward-propagation process. This adjustment refines the forecasting outputs by compensating for autocorrelated errors, thereby improving prediction accuracy.

The SAEA framework is model-agnostic, meaning it can be seamlessly integrated with existing state-of-the-art traffic forecasting models without requiring modifications to their architectures. By systematically addressing spatiotemporal error correlations, SAEA enhances the robustness and reliability of DNN-based forecasting systems.

### 3.1. Vector autoregressive process in prediction errors

A one-step-ahead traffic forecasting model is typically formulated as:

$$\mathcal{G}_t = f\left(\mathcal{G}_{t-1}, \ldots, \mathcal{G}_{t-H}; \theta\right) + \epsilon_t, \quad (5)$$

where $\epsilon_t$ represents the prediction error. The model is often trained by minimizing the loss functions MSE $\sim \sum_t \|\epsilon_t\|_2$ and MAE $\sim \sum_t \|\epsilon_t\|_1$, which correspond to independent Gaussian and independent Laplacian noise assumptions, respectively.

Empirical results, as shown in Fig. 1, indicate that the independent noise assumption does not hold in real-world traffic forecasting. To account for spatiotemporal autocorrelation, we redefine the error term $\epsilon_t$ in Eq. (5) as $\eta_t$:

$$\mathcal{G}_t = f\left(\mathcal{G}_{t-1}, \ldots, \mathcal{G}_{t-H}; \theta\right) + \eta_t, \quad (6)$$

where $\eta_t$ follows a vector autoregressive process VAR($p$):

$$\eta_t = \Phi_1 \eta_{t-1} + \cdots + \Phi_p \eta_{t-p} + \epsilon_t, \quad (7)$$

where $\epsilon_t \sim N(\mathbf{0}, \Sigma)$ is a Gaussian white noise process, and $\Phi_1, \ldots, \Phi_p$ are coefficient matrices of size $N \times N$.

This formulation represents the prediction error $\eta_t$ as a VAR process. However, existing traffic forecasting models typically treat prediction errors as i.i.d., disregarding the inherent spatiotemporal autocorrelations. To address this limitation, it is essential to explicitly model the VAR process of prediction errors.

### 3.2. New cost function based on spatiotemporal correlation fusion

To adjust for autocorrelated errors, we employ a VAR(1) model in DNN-based traffic forecasting. By combining Eqs. (6) and (7), the updated traffic forecasting model is formulated as:

$$\mathcal{G}_t = f\left(\mathcal{G}_{t-1}, \ldots, \mathcal{G}_{t-H}; \theta\right) + \Phi \eta_{t-1} + \epsilon_t, \quad (8)$$

where $\eta_{t-1}$, the historical prediction error, is computed as:

$$\eta_{t-1} = \mathcal{G}_{t-1} - f\left(\mathcal{G}_{t-2}, \ldots, \mathcal{G}_{t-H-1}; \theta\right). \quad (9)$$

Replacing $\eta_{t-1}$ in Eq. (8) with Eq. (9), we reformulate the traffic forecasting model as:

$$\mathcal{G}_t - \Phi \mathcal{G}_{t-1} = f(\mathcal{G}_{t-1}, \ldots, \mathcal{G}_{t-H}; \theta) - \Phi f(\mathcal{G}_{t-2}, \ldots, \mathcal{G}_{t-H-1}; \theta) + \epsilon_t. \quad (10)$$

This new formulation accounts for autocorrelated errors, but its complexity poses challenges in direct estimation. To simplify, we approximate the right-hand side of Eq. (10) as:

$$\mathcal{G}_t - \Phi \mathcal{G}_{t-1} = f(\mathcal{G}_{t-1} - \Phi \mathcal{G}_{t-2}, \ldots, \mathcal{G}_{t-H} - \Phi \mathcal{G}_{t-H-1}; \theta) + \epsilon_t. \quad (11)$$

This transformation ensures structural consistency between the left and right sides, maintaining the same input–output form while allowing parameter estimation using a standard deep neural network. The parameters $\Phi$ and $\theta$ are jointly learned, allowing integration with state-of-the-art forecasting models. To ensure fair evaluation and avoid extending the historical window, we replace $\mathcal{G}_{t-H-1}$ with its empirical mean:

$$\mathcal{G}_{t-H-1} = \frac{1}{H} \sum_{h=1}^{H} \mathcal{G}_{t-h}. \quad (12)$$

The model parameters, along with the coefficient matrix $\Phi$, are optimized using Stochastic Gradient Descent (SGD). Since $\Phi$ must maintain stationarity, we introduce an $\ell_1$-norm regularization term, $\mathcal{R} = \|\Phi\|_1$, to enforce sparsity.

The final cost function used for training is:

$$loss = \left\| \mathcal{G}_t - \Phi \mathcal{G}_{t-1} - f(\mathcal{G}_{t-1} - \Phi \mathcal{G}_{t-2}, \ldots, ; \theta) \right\|_2 + \alpha \cdot \mathcal{R}, \quad (13)$$

where $\alpha > 0$ is a penalty coefficient. Training with the new cost function, we can directly learn both the model parameter and the coefficient matrix.

The inference process is reformulated as an autoregressive framework, where the learned coefficient matrix is utilized to adjust prediction errors in forecasting models. The one-step-ahead prediction is computed as:

$$\mathcal{G}_t^* = \Phi \mathcal{G}_{t-1} + f(\mathcal{G}_{t-1} - \Phi \mathcal{G}_{t-2}, \ldots, ; \theta). \quad (14)$$

### 3.3. Multi-step extension

The proposed cost function in Eq. (13) can be effectively applied to train recursive multi-step forecasting models, as formulated in Eq. (3). However, recursively generating one-step-ahead predictions often leads to significant error accumulation, particularly for long-term forecasting [33].

To mitigate this issue and enhance forecasting accuracy, we adopt a multi-step forecasting approach using multiple models, as formulated in Eq. (4), which allows each forecasting step to be learned independently. The autocorrelated error can be adjusted by defining loss based on $\epsilon_t$ like Eq. (8):

$$\mathcal{G}_{t+p} = f_p(\mathcal{G}_{t-1}, \ldots, \mathcal{G}_{t-H}; \theta_p) + \Phi_p \eta_{t-1} + \epsilon_t, \quad (15)$$

where $f_p(\cdot)$ denotes the forecasting model used to predict traffic state at step-$p$ and $\Phi_p$ denotes the coefficient matrix capturing the error correlation for the corresponding forecasting model.

The cost function based on Eq. (15) is defined as:

$$loss = \left\| \mathcal{G}_{t+p} - \Phi_p \mathcal{G}_{t-1} - f_p(\mathcal{G}_{t-1} - \Phi_p \mathcal{G}_{t-2}, \ldots, ; \theta) \right\|_2 + \alpha \cdot \mathcal{R}, \quad (16)$$

where $\mathcal{R} = \|\Phi_p\|_1$ denotes the $\ell_1$-norm regularization. This formulation adjusts the forecast by introducing a correction term that accounts for autocorrelated errors via $\Phi_p$. Importantly, both the model parameters $\theta$ and the error adjustment matrix $\Phi_p$ are jointly optimized during training by minimizing the loss $loss$. Since $\mathcal{G}_{t-1}, \mathcal{G}_{t-2}, \ldots$ are available historical inputs and $\mathcal{G}_{t+p}$ is observed ground truth, the residuals can be computed directly, enabling the model to learn both $\theta$ and $\Phi_p$ in a fully supervised manner.

The step-$p$ forecasting in the running process can be obtained by

$$\mathcal{G}_t^* = \Phi_p \mathcal{G}_{t-1} + f_p(\mathcal{G}_{t-1} - \Phi_p \mathcal{G}_{t-2}, \ldots, ; \theta). \quad (17)$$

Lastly, it should be noted that SAEA is not applicable to Seq2Seq-based forecasting models (Eq. (2)), as their autoregressive structure fundamentally differs from the direct forecasting paradigm.

### 3.4. Spatial structure embedding

While modeling the prediction error as a vector autoregressive (VAR) process effectively captures temporal dependencies, it does not account for the inherent spatial correlations present in traffic data. In particular, prior structural information, such as road network topology and sensor connectivity, is not explicitly incorporated into the cost functions defined in Eqs. (13) and (16).

To address this limitation, we introduce a structurally sparse regularization that embeds prior spatial information into the coefficient matrix, enhancing its ability to model spatial dependencies in traffic data.

Supposing $D$ represents the degree matrix and $W$ is the adjacency matrix capturing spatial relationships between different sensors, a spatial mask, $M$, designed to enforce structural sparsity, is computed as:

$$M = 1 - \left\lceil \left\| \widetilde{L} \right\| \right\rceil, \quad (18)$$

where the operator $\|\cdot\|$ computes the absolute value while $\lceil \cdot \rceil$ represents the ceiling operation. Here, $\widetilde{L}$ is the normalized Laplacian matrix, which is derived from the Laplacian matrix: $L = D - W$.

Along with the spatial mask, the proposed structurally sparse regularization term is formulated as:

$$\mathcal{R} = \left\| M \odot \Phi_p \right\|_2, \quad (19)$$

where $\odot$ denotes the Hadamard product.

By incorporating this structured regularization, our approach ensures that the coefficient matrix faithfully reflects the inherent spatial dependencies within the traffic network. This enhancement not only strengthens forecasting performance by effectively modeling localized spatial correlations but also improves the interpretability of the learned relationships, providing deeper insights into the underlying traffic dynamics.

The proposed structural regularization incorporates a mask, as illustrated in Fig. 3(g), to enforce structural sparsity in the coefficient matrix. By integrating this regularization term, the model effectively captures spatial dependencies among prediction errors, as demonstrated in Fig. 3(e), providing interpretability regarding how errors propagate across neighboring nodes. Fig. 3(h) presents the second-order mask, which is derived from the second-order adjacency matrix, while Fig. 3(f) shows the corresponding learned coefficient matrix. Compared to the first-order structurally sparse coefficient matrix, the second-order version incorporates information from more distant neighbors, enabling a broader exploration of error correlations within the network.

### 3.5. Possible regularization term

Initially, the coefficient matrix is regularized using an $\ell_1$-norm sparsity constraint, as defined in Eq. (13). Subsequently, it is refined to enforce structural sparsity. This section explores alternative regularization strategies that can serve as baseline comparisons for evaluating the effectiveness of the proposed spatial structure embedding.

**Temporal Scalar.** The coefficient matrix, $\Phi_p$, can be replaced as a scalar, $\mu_p$, which only accounts for the temporally autocorrelated errors, and all variates share the same coefficient. This is a simple extension of the work adjusting autocorrelated errors in univariate forecasting [12]. The scalar's absolute value should be less than 1 and thus the regularization term is

$$\mathcal{R} = \max\left(0, \left\| \mu_p \right\| - 1\right). \quad (20)$$

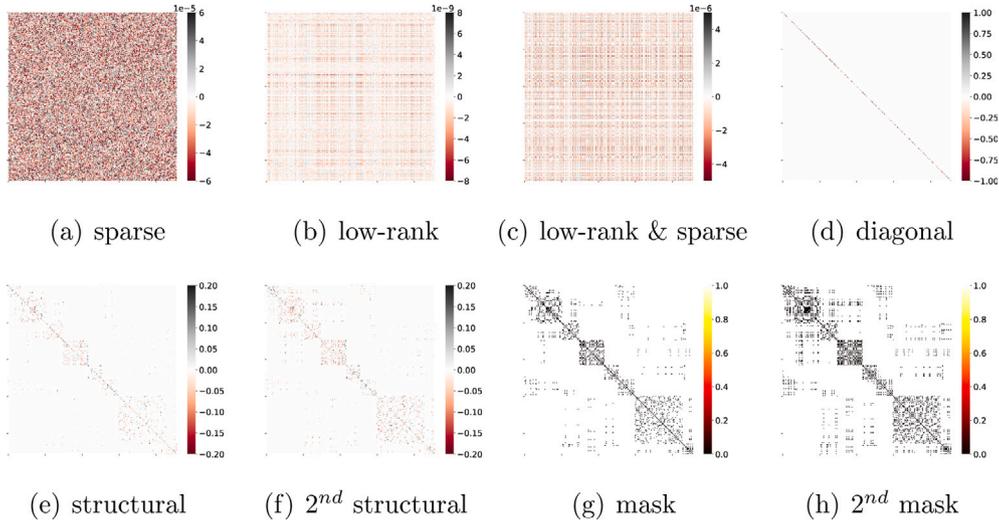

(a) sparse  (b) low-rank  (c) low-rank & sparse  (d) diagonal

(e) structural  (f) $2^{nd}$ structural  (g) mask  (h) $2^{nd}$ mask

**Fig. 3.** Comparison of coefficient matrices, $\Phi_p$, learned under different regularization strategies. (a) shows the $\ell_1$-norm sparse coefficient matrix, (b) shows the low-rank coefficient matrix, and (c) shows the coefficient matrix with low-rank & sparse regularization. (d) shows a purely diagonal coefficient matrix, where only self-dependencies are retained. (e) shows the proposed structurally sparse coefficient matrix from the mask (g), and (f) shows the structurally sparse coefficient matrix learned from a second-order mask (h). The proposed structurally sparse regularization not only enforces sparsity but also captures spatial relationships, improving interpretability by revealing how prediction errors propagate across neighboring nodes.

**Diagonal Matrix.** The coefficient matrix can be forced into a diagonal matrix, $\Phi_p = diag(\mathbf{v_p})$, where $\mathbf{v_p} = [v_1^p, \ldots, v_N^p]$ and $N$ denotes the number of variates. The diagonal coefficient matrix only accounts for the temporally autocorrelated errors but entries are different. Similarly, the absolute value of ever entry in the diagonal matrix should be less than 1 and the regularization term is

$$\mathcal{R} = \sum_{i=1}^{N} \max\left(0, \left\|v_i^p\right\| - 1\right). \tag{21}$$

**Low-Rank Matrix.** The coefficient matrix is reformed into a low-rank matrix, which can be represented as

$$\Phi_p = P_p Q_p, \tag{22}$$

where the matrix is decomposed into an $N \times k$ matrix $P_p$ and a $k \times N$ matrix $Q_p$, and $k \ll N$. The regularization term is

$$\mathcal{R} = \left\|P_p\right\|_2 + \left\|Q_p\right\|_2. \tag{23}$$

**Low-Rank and Sparse Matrix.** We can integrate low-rank decomposition with a sparsity constraint to enhance the coefficient matrix's structure. Specifically, the coefficient matrix is decomposed into a low-rank component and a sparse component, formulated as:

$$\Phi_p = P_p Q_p + S_p, \tag{24}$$

where $P_p Q_p$ captures the global dependencies and smooth variations in the spatiotemporal data, while $S_p$ preserves localized and abrupt changes. This decomposition effectively balances global structure learning with fine-grained adaptiveness, improving interpretability and robustness in traffic forecasting. The modified regularization term is

$$\mathcal{R} = \left\|P_p\right\|_2 + \left\|Q_p\right\|_2 + \frac{\beta}{\alpha} \cdot \left\|S_p\right\|_1, \tag{25}$$

where $\beta$ is the hyperparameter to balance the low-rank form and the sparse form.

A toy empirical study, illustrated in Fig. 3, is conducted to explicitly compare different parameterization strategies. This experiment applies a 45-minute-ahead prediction using STGCN on the PeMS dataset. The coefficient matrices obtained from different regularization techniques reveal distinct structural properties. As shown in Figs. 3(a)–3(c), it is challenging to discern meaningful correlation patterns since the matrix entries are either highly similar or close to zero, limiting their interpretability. The diagonal matrix in Fig. 3(d) neglects spatial dependencies, making it insufficient for capturing complex error correlations in traffic forecasting. In contrast, the structurally sparse coefficient matrix, illustrated in Fig. 3(e), exhibits a well-defined structural form. A significant portion of its elements are exactly zero, while the remaining nonzero elements have reasonably large absolute values. This characteristic enhances interpretability, as it clearly reveals how the prediction error at one vertex is influenced by neighboring nodes, thereby improving the model's ability to account for spatial dependencies in forecasting tasks.

### 3.6. Insightful discussion

Designing a forecasting model that perfectly eliminates autocorrelated errors is theoretically possible; however, such an approach lacks generalizability. A model that precisely captures hidden patterns in one dataset is likely to become misspecified when applied to other datasets, limiting its practical applicability.

In this work, traffic prediction errors are treated as autocorrelated across both spatial and temporal dimensions. Instead of assuming a white noise process, we model these errors using a Vector Autoregressive (VAR) process. To account for spatiotemporal dependencies, we introduce a coefficient matrix that is embedded into the cost function and jointly learned with the model parameters. As defined in Eq. (14), the forecasting output consists of two components: the primary model's prediction $f(\mathcal{G}_{t-1} - \Phi\mathcal{G}_{t-2}, \ldots; \theta)$ and the error adjustment term $\Phi\mathcal{G}_{t-1}$. Based on the VAR assumption, prediction errors are modeled as being linearly correlated with the most recent traffic states. The Pseudocode of the training process of SAEA is shown as Algorithm 1.

When the coefficient matrix is a zero matrix, the updated cost function in Eq. (13) reduces to the conventional Mean Squared Error (MSE) minimization, implicitly assuming that prediction errors are independent and identically distributed (i.i.d.). This ensures that models trained with the updated cost function perform at least as well as those trained with the original MSE-based loss. A similar principle, where introducing an external component does not degrade performance, is commonly employed in residual connections [34–37]. However, unlike these prior works, which focus on modifying the internal architecture of DNNs, our approach reformulates the cost function itself, enhancing the model's ability to account for structured spatiotemporal dependencies.

**Algorithm 1** Spatiotemporal Autocorrelated Error Adjustment

1: **Input:** the traffic forecasting model $f(\cdot)$, historical horizon $H$, traffic observations $\mathcal{G}_t$, coefficient matrix $\Phi$, and mask $M$.
2: **Output:** training loss $loss$, and single-step prediction $\mathcal{G}_t^*$.
3: **while** not converged **do**
4: $\quad \mathcal{G}_t' = [\mathcal{G}_{t-1}, \mathcal{G}_{t-2}, ...\mathcal{G}_{t-H}] - \Phi \left[\mathcal{G}_{t-2}, \mathcal{G}_{t-3}, ...\mathcal{G}_{t-H}, \frac{1}{H}\sum_{h=1}^{H}\mathcal{G}_{t-h}\right]$
5: $\quad bias = f(\mathcal{G}_t')$
6: $\quad \mathcal{G}_t^* = bias + \Phi\mathcal{G}_{t-1}$
7: $\quad loss = \ell_2(\mathcal{G}_t - \mathcal{G}_t^*) + 10 \cdot \ell_2(M \odot \Phi)$
8: **end while**
9: Return the training loss and sing-step prediction.

In summary, our proposed SAEA framework introduces several key innovations over traditional forecasting methods. First, instead of assuming independent and identically distributed (i.i.d.) errors, SAEA models the residuals as a vector autoregressive (VAR) process, capturing both temporal and spatial correlations in prediction errors. Second, this leads to a novel training objective that adjusts the forecasting model to account for autocorrelated errors, replacing standard MSE loss with a theoretically grounded, VAR-informed cost function. Third, we introduce a structured sparsity constraint on the coefficient matrix, which not only enhances training efficiency but also improves interpretability by aligning the learned error propagation patterns with the underlying spatial topology. Collectively, these advances redefine the training process of deep forecasting models by explicitly addressing the limitations of the i.i.d. assumption.

## 4. Experiments

The proposed SAEA is validated across a wide range of traffic forecasting models on different traffic data to prove its effectiveness and model-agnostic.

### 4.1. Models

Our experiments evaluate the proposed approach using five traffic forecasting models, encompassing both traditional statistical methods and state-of-the-art deep learning-based models. The selected models are as follows:

- **SVR**: Support Vector Regression (SVR) [38], a classical machine learning model for time series regression.
- **FC-LSTM**: Fully Connected Long Short-Term Memory (FC-LSTM) [39], which captures long-range temporal dependencies in sequential data.
- **STGCN**: Spatiotemporal Graph Convolutional Network (STGCN) [5], which combines spatial GNN layers and temporal gated-CNN layers, integrating spatial and temporal dependencies.
- **GWave**: Graph WaveNet [6], which employs dilated causal convolutions and adaptive graph learning for traffic forecasting.
- **AutoSTG**: Automated Spatiotemporal Graph Prediction (AutoSTG) [40], a neural architecture search-based model designed to optimize spatiotemporal graph structures.

The experiments are conducted on an Ubuntu 18.04 LTS system with TensorFlow 1.18, Python 3.7.4, and a Tesla V100 GPU. All hyperparameters, except for the input and output channels, are configured according to their respective original papers to ensure a fair comparison. In all models, the input sequence length is set to 12, meaning that 12 historical observations are fed into the forecasting models. The output prediction is set to a single time step per model, allowing us to train multiple models for different forecasting horizons as formulated in Eq. (4).

### 4.2. Datasets

The datasets used in these experiments are PeMS (Freeway Performance Measurement System) and METR-LA. PeMS is collated by the California Department of Transportation (Caltrans). Data collected from 228 road segments and 44 days are used in these experiments, which is as same as STGCN [5]. METR-LA contains highway information in Los Angeles County. Data from 207 sensors and 4 months are applied in our experiments, which is as same as DCRNN [4].

### 4.3. Metrics

To assess the prediction performance of the proposed models, we employ three widely used evaluation metrics [41–43]:

- **Mean Absolute Percentage Error (MAPE)**: MAPE measures the average percentage deviation between predicted and actual values, providing an interpretable relative error metric. It is defined as:

$$\text{MAPE} = \frac{1}{N}\sum_{i=1}^{N}\left|\frac{Y_i - \hat{Y}_i}{Y_i}\right| \times 100\%, \tag{26}$$

where $Y_i$ and $\hat{Y}_i$ denote the ground truth and predicted values, respectively.

- **Root Mean Square Error (RMSE)**: RMSE quantifies the standard deviation of residuals, emphasizing larger errors by squaring deviations before averaging. It is given by:

$$\text{RMSE} = \sqrt{\frac{1}{N}\sum_{i=1}^{N}(Y_i - \hat{Y}_i)^2}. \tag{27}$$

- **Error Correlation Matrix (ECM)**: To capture spatial and temporal dependencies in prediction errors, we introduce the error correlation matrix as an additional evaluation metric. The ECM quantifies how prediction errors at different sensors (spatial correlation) or different time steps (temporal correlation) are interrelated. It is computed as:

$$\mathbf{ECM} = \frac{\mathbf{E}\mathbf{E}^T}{N}, \tag{28}$$

where $\mathbf{E}$ is the matrix of residuals, with each row representing the error vector at a given time step across all sensors.

These metrics collectively provide a comprehensive evaluation, with MAPE offering a percentage-based error measure, RMSE highlighting absolute deviations, and ECM revealing spatial and temporal correlations in forecasting errors.

### 4.4. Baselines

In addition to the straightforward extension of temporal scalars used in existing autocorrelated error adjustment approaches for univariate forecasting [12], we incorporate a variety of parameterization strategies for comparison. These include:

- $\ell_1$-**norm sparse**: Enforces sparsity in the coefficient matrix by minimizing the $\ell_1$-norm, ensuring that only a few elements contribute significantly to the error adjustment.
- **Diagonal**: Restricts the coefficient matrix to a diagonal form, assuming that each time series' error is only dependent on its past values without considering interdependencies.
- **Low-rank**: Decomposes the coefficient matrix into a low-rank approximation, capturing dominant global structures while discarding minor variations.
- **Low-rank & sparse**: Combines low-rank decomposition with sparse constraints to balance global structure representation and local sparsity.

**Table 1**
Overall comparison.

| Model | PeMS (15/30/45 min) | | META-LA (15/30/45 min) | |
|---|---|---|---|---|
| | MAPE (%) | RMSE | MAPE (%) | RMSE |
| SVR(no adjustment) | 5.81/8.88/11.50 | 4.55/6.67/8.28 | 8.33/9.48/12.42 | 6.25/7.10/8.72 |
| SVR [12] | 5.67/8.73/11.25 | 4.48/6.59/8.07 | 8.27/9.35/12.32 | 6.22/7.01/8.48 |
| SVR(diagonal) | 5.64/8.72/11.24 | 4.46/6.42/7.44 | 8.28/9.33/12.09 | 6.20/6.82/8.36 |
| SVR(sparse) | 5.61/**8.70**/11.22 | 4.46/6.50/7.39 | 8.35/9.28/12.22 | 6.13/6.84/8.11 |
| SVR(low-rank) | 5.63/8.74/11.30 | 4.46/6.37/7.40 | 8.18/9.34/12.28 | 6.16/6.93/8.14 |
| SVR(low-rank+sparse) | 5.61/8.71/**11.17** | 4.45/6.42/7.28 | **8.15**/9.31/12.24 | 6.14/6.90/8.49 |
| SVR(**proposed**) | **5.60**/8.72/11.23 | **4.43**/**6.35**/**7.22** | 8.25/**9.26**/**12.07** | **6.12**/**6.77**/**8.05** |
| FC-LSTM(no adjustment) | 8.60/9.55/10.10 | 6.20/7.03/7.51 | 11.10/11.41/11.69 | 7.68/7.94/8.20 |
| FC-LSTM [12] | 8.55/9.51/9.91 | 6.14/6.91/7.29 | 11.08/11.38/11.50 | 7.69/7.88/8.14 |
| FC-LSTM(diagonal) | 8.50/9.49/**9.87** | 6.04/6.85/7.15 | 11.05/11.36/11.51 | 7.66/7.87/7.98 |
| FC-LSTM(sparse) | **8.46**/9.42/9.94 | 6.02/6.79/**7.01** | 11.03/11.35/11.48 | 7.65/7.82/8.01 |
| FC-LSTM(low-rank) | 8.51/9.46/9.91 | 6.05/6.87/7.16 | 11.04/11.37/11.56 | 7.64/7.85/7.97 |
| FC-LSTM(low-rank+sparse) | 8.47/9.43/9.99 | **6.00**/6.81/7.35 | **11.01**/**11.33**/11.52 | 7.60/7.82/8.09 |
| FC-LSTM(**proposed**) | 8.57/**9.41**/9.89 | **6.00**/**6.78**/7.03 | 11.05/11.34/**11.44** | **7.59**/**7.80**/**7.91** |
| STGCN(no adjustment) | 5.24/7.39/9.12 | 4.06/5.74/7.01 | 7.62/9.57/10.12 | 5.73/7.22/7.78 |
| STGCN [12] | 5.24/7.37/9.10 | 4.08/5.70/6.75 | **7.60**/9.44/10.06 | 5.71/7.09/7.72 |
| STGCN(diagonal) | 5.23/7.36/9.04 | 4.07/5.74/6.58 | **7.60**/9.57/10.04 | 5.69/6.98/7.70 |
| STGCN(sparse) | **5.22**/7.35/**8.49** | 4.05/5.50/6.33 | 7.63/9.48/**10.02** | 5.66/6.94/7.68 |
| STGCN(low-rank) | 5.27/**7.34**/8.79 | 4.06/5.59/6.48 | 7.62/9.46/10.03 | 5.67/6.99/7.65 |
| STGCN(low-rank+sparse) | 5.39/7.40/9.14 | 4.05/5.61/6.38 | 7.62/**9.44**/10.05 | 5.66/6.95/7.72 |
| STGCN(**proposed**) | 5.29/7.37/8.96 | **4.02**/**5.46**/**6.26** | 7.64/9.47/10.07 | **5.64**/**6.92**/**7.61** |
| AutoSTG(no adjustment) | 5.22/7.28/8.65 | 4.03/5.66/6.71 | 6.94/8.40/9.59 | 5.16/6.17/7.18 |
| AutoSTG [12] | 5.31/7.30/8.61 | 4.02/5.61/6.48 | 6.93/8.35/9.55 | 5.19/6.19/7.12 |
| AutoSTG(diagonal) | 5.22/7.25/8.57 | 4.00/5.56/6.59 | 7.04/**8.31**/9.57 | 5.20/6.23/7.10 |
| AutoSTG(sparse) | 5.22/7.25/8.61 | 4.04/5.58/6.30 | 6.99/8.37/**9.50** | 5.15/6.18/7.11 |
| AutoSTG(low-rank) | 5.28/7.26/8.62 | 4.02/5.58/6.47 | 6.96/8.39/9.51 | 5.16/**6.16**/7.09 |
| AutoSTG(low-rank+sparse) | 5.21/7.24/8.61 | 4.01/5.59/6.24 | 7.13/8.34/9.54 | 5.15/6.20/7.13 |
| AutoSTG(**proposed**) | **5.20**/**7.23**/**8.54** | **3.98**/**5.57**/6.33 | **6.90**/8.35/9.60 | **5.14**/**6.16**/**7.01** |
| GWave(no adjustment) | 5.21/7.67/8.66 | 4.13/5.78/6.92 | 6.90/8.37/9.68 | 5.14/6.20/7.42 |
| GWave [12] | 5.22/7.60/8.61 | 4.14/5.73/6.48 | 6.86/8.35/9.62 | 5.16/6.19/7.30 |
| GWave(diagonal) | **5.20**/7.58/8.60 | 4.13/5.80/6.90 | 6.84/8.35/9.60 | 5.13/6.18/7.35 |
| GWave(sparse) | 5.22/7.57/**8.56** | 4.16/5.74/6.40 | 6.88/8.36/9.63 | 5.14/6.17/7.19 |
| GWave(low-rank) | 5.23/7.57/8.59 | 4.14/5.73/6.31 | 6.90/8.34/9.58 | 5.15/6.17/7.21 |
| GWave(low-rank+sparse) | **5.20**/7.55/8.57 | 4.13/5.83/6.51 | **6.83**/8.41/**9.57** | 5.14/6.22/7.28 |
| GWave(**proposed**) | 5.24/7.61/8.55 | **4.11**/**5.71**/**6.24** | 6.84/**8.33**/9.58 | **5.12**/**6.14**/**7.15** |

- **Proposed structurally sparse**: Embeds prior spatial information to enforce structured sparsity, effectively modeling spatial dependencies in traffic data.

By systematically evaluating these parameterizations, we aim to assess their effectiveness in capturing autocorrelated errors and improving forecasting robustness.

### 4.5. Overall comparison

All traffic forecasting models were evaluated on the PeMS and META-LA datasets. For all experiments, the number of epochs was set to 300, the learning rate was fixed at $5e^{-4}$, the batch size was 50, and the RMSProp optimizer was used. The results for 15-min, 30-min, and 45-minute-ahead predictions are recorded.

Table 1 presents the prediction performance of various traffic forecasting models, comparing their original implementations to the same models adjusted with the proposed spatiotemporal autocorrelated error adjustment (SAEA) framework. Each model is tested with multiple parameterizations. Taking "SVR" as an example, "SVR (no adjustment)" represents the baseline model without any error adjustment, while "SVR [12]" applies a temporal scalar adjustment as a simple extension of [12]. Additionally, "SVR (diagonal)" adjusts errors using a diagonal coefficient matrix, "SVR (sparse)" applies $\ell_1$-norm sparsity, "SVR (low-rank)" uses low-rank decomposition, and "SVR (low-rank + sparse)" combines low-rank and sparse constraints. Finally, "SVR (**proposed**)" applies the structurally sparse coefficient matrix as defined in Eq. (19), representing the complete SAEA approach. The best-performing model in each category is highlighted in **bold**.

From Table 1, it is evident that the proposed SAEA framework consistently enhances the predictive performance of both traditional and deep learning-based traffic forecasting models across different parameterizations. Notably, SAEA encoding prior spatial information outperforms all other configurations in most cases. The improvements are particularly pronounced for long-term forecasting (i.e., 45-min predictions), where SAEA significantly reduces errors. While the improvements in MAPE might appear marginal, the reductions in RMSE are substantial. For instance, in nearly all models, 45-minute-ahead RMSE values decrease by 5% to over 10% when applying SAEA. These results quantitatively validate the effectiveness and generalization of SAEA.

An interesting observation is that the performance gains for AutoSTG on the META-LA dataset are less pronounced than in other cases. This is related to model misspecification and the degree to which correlated errors are already captured by the base model. In theory, if a forecasting model could perfectly capture the structure of correlated errors, there would be little room for further improvement–however, this is rarely achievable in practice. SAEA is designed to bridge the gap between a model's current performance and its ideal.

In specific cases, such as AutoSTG on the META-LA dataset, the base model may already be highly optimized and capable of modeling correlations effectively. As a result, the relative performance gain from SAEA appears smaller. Nevertheless, SAEA still provides a systematic and model-agnostic enhancement and can be especially valuable in cases where the base model underfits or partially captures structured errors.

### 4.6. Interpretability

Fig. 4 presents the 30-min and 45-minute-ahead traffic forecasting results and corresponding prediction errors for STGCN with and without the proposed SAEA on the PeMS dataset. The green-shaded area represents the prediction errors in the original STGCN model, while the purple-shaded area indicates errors after applying SAEA. The results clearly demonstrate that the incorporation of SAEA significantly reduces prediction errors, thereby enhancing the accuracy and robustness of the forecasting model. These findings strongly validate the effectiveness of SAEA in improving the performance of existing traffic forecasting frameworks by mitigating autocorrelated prediction errors.

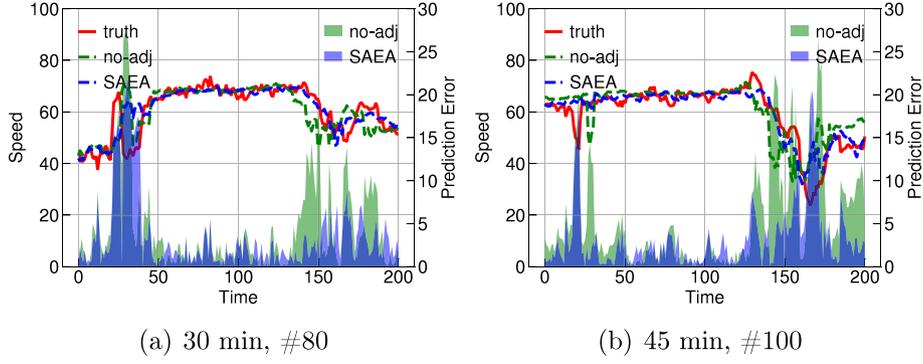

**Fig. 4.** Prediction errors with and without SAEA. Panels (a) and (b) present a comparison of 30-min and 45-minute-ahead predictions on nodes #80 and #100, respectively. The solid red line represents the ground truth, while the dashed green line ("no-adj") corresponds to predictions by STGCN without autocorrelated error adjustment. The dashed blue line ("SAEA") shows predictions with the proposed spatiotemporal autocorrelated error adjustment. The shaded areas indicate the magnitude of prediction errors. The results demonstrate that SAEA effectively reduces prediction errors by accounting for spatiotemporal dependencies in forecasting models.

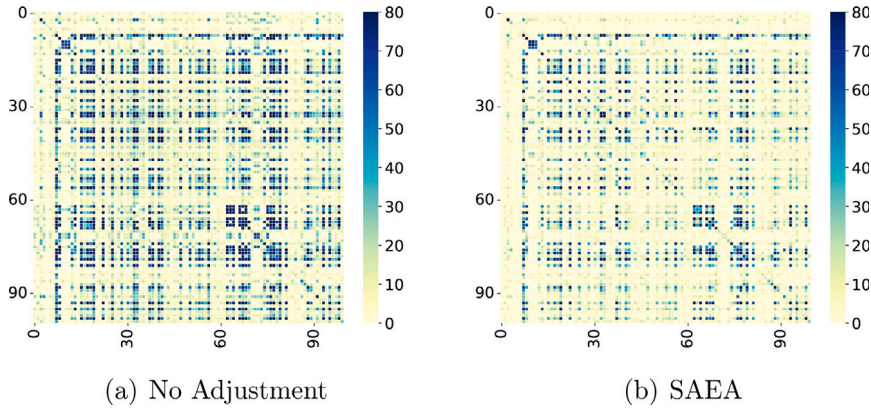

**Fig. 5.** Spatiotemporal autocorrelation of prediction errors for 45-minute-ahead forecasting on PeMS. Panels (a) and (b) illustrate the covariance matrices of prediction errors from STGCN without adjustment and with the proposed SAEA, respectively. The comparison highlights that SAEA effectively reduces spatiotemporal error correlations, demonstrating its capability to enhance forecasting accuracy.

Fig. 5 presents the spatiotemporal correlation of 45-minute-ahead prediction errors from STGCN with and without the proposed SAEA on the PeMS dataset. Fig. 5(a), which illustrates the error covariance matrix of the original model, appears significantly more cluttered compared to Fig. 5(b), where SAEA has been applied. This comparison clearly demonstrates that the proposed SAEA effectively accounts for and mitigates autocorrelated errors, thereby enhancing the overall forecasting accuracy.

Fig. 6 presents the structurally sparse coefficients for different nodes, forecasting horizons combination, illustrating the spatial correlation of prediction errors on different nodes. Higher absolute coefficient values indicate stronger error correlations between nodes. For example, the errors at node #10 exhibit strong dependencies on itself and node #6, while the errors at node #100 are significantly influenced by nodes #87 and #89. This structural pattern highlights the spatiotemporal relationships captured by the proposed SAEA framework, reinforcing its ability to model and adjust autocorrelated prediction errors effectively.

### 4.7. Hyperparameter study

A series of experiments are conducted to identify the optimal hyperparameters for different parameterizations. The selected potential hyperparameters are as follows.

- For parameterizations such as $\ell_1$-norm sparse, temporal scalar [12], diagonal, and structurally sparse, the penalty coefficient, $\alpha$, is selected from the set $\{1e^0, 1e^1, 1e^2, 1e^3, 1e^4, 1e^5\}$.
- For the low-rank parameterization, the hyperparameters $k$ and $\alpha$ are chosen from the sets $\{5, 10, 15, 20, 25\}$ and $\{1e1, 1e2, 1e3, 1e4\}$, respectively.
- For the low-rank & sparse parameterization, the hyperparameters $k$, $\alpha$, and $\beta$ are selected from the sets $\{5, 10, 15, 20, 25\}$, $\{1e1, 1e2, 1e3\}$, and $\{1e1, 1e2, 1e3\}$, respectively.

45 minute-ahead predictions by STGCN [5] on PeMS are recorded and RMSE works as the measurement. Comparison results are shown in Fig. 7. Based on this experiment, hyperparameters are set as Table 2.

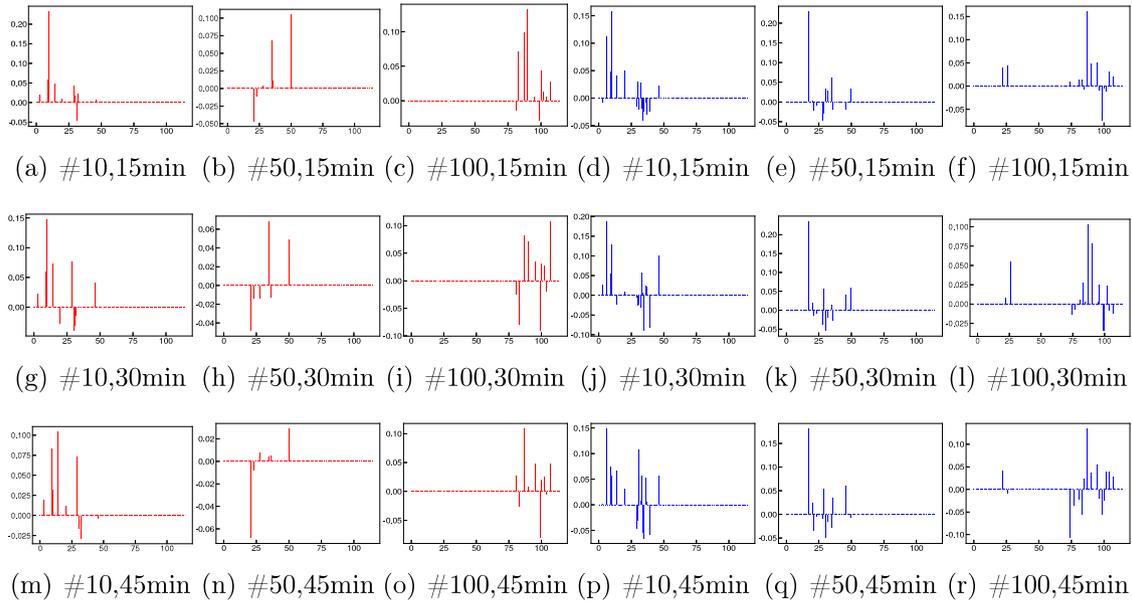

**Fig. 6.** Structural Coefficients with STGCN on PeMS Data. Each subfigure illustrates the learned coefficient values for different nodes and forecasting horizons. "Red" represents the $1st$-order structurally sparse coefficients, while "blue" represents the $2nd$-order structurally sparse coefficients. The notation "#10,15 min" represents the coefficients for the 15-minute-ahead prediction at node #10. Columns from left to right correspond to node #10, #50, and #100, respectively for every color; rows from top to bottom correspond to 15-min, 30-min, and 45-minute-ahead predictions, respectively. The $x$-axis represents the node index, ranging from 0 to 120. Notably, the coefficients are all zero for indices from 121 to 228, indicating a limited spatial influence beyond the observed range.

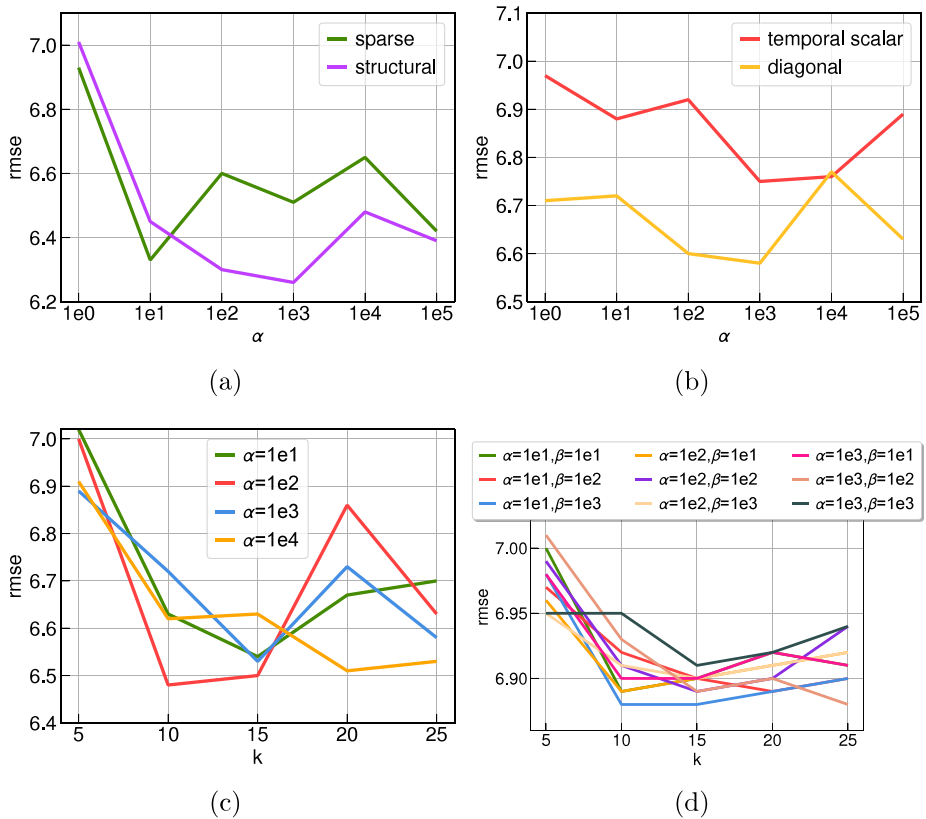

**Fig. 7.** Hyperparameter Comparison. Panel (a) presents the hyperparameter study of two parameterizations — $\ell_1$-norm sparse and the proposed structurally sparse — showing the effect of different penalty coefficients ($\alpha$) on RMSE. Panel (b) evaluates the impact of $\alpha$ on temporal scalar and diagonal parameterizations. Panel (c) investigates the effect of the rank parameter ($k$) and $\alpha$ on the low-rank parameterization, while Panel (d) examines the combined influence of $k$, $\alpha$, and $\beta$ on the low-rank & sparse parameterization.

**Table 2**
Hyperparameter settings.

| | |
|---|---|
| Sparse | $\alpha = 100$ |
| Temporal scalar | $\alpha = 1000$ |
| Diagonal | $\alpha = 1000$ |
| Low-rank | $k = 10$ |
| | $\alpha = 100$ |
| Low-rank & Sparse | $k = 10$ |
| | $\alpha = 10$ |
| | $\beta = 1000$ |
| Structural | $\alpha = 1000$ |

*4.8. Order analysis in VAR*

In Eq. (8), VAR(1) rather than higher-order VAR($p$) is applied, because of the trade-off between theoretical expressiveness and practical feasibility. While a VAR($p$) model with $p > 1$ can theoretically capture more complex and longer-range temporal autocorrelation in the error sequence, in practice, increasing $p$ significantly enlarges the parameter space, making the training more challenging when combined with deep neural networks in traffic forecasting.

To explore this further, we compared the error adjustment performance using both VAR(1) and VAR(2) models. The results are shown in Table 3. Interestingly, despite the theoretical advantage of higher-order VAR models, we observe that VAR(1) often outperforms VAR(2) in practice. This suggests that the added complexity of modeling longer-range dependencies may introduce optimization and convergence difficulties, leading to worse generalization.

The experimental results consistently demonstrate the effectiveness of the proposed SAEA framework in improving traffic forecasting accuracy across multiple models and datasets. The quantitative evaluation, presented in Table 1, reveals that models integrated with SAEA outperform their baseline counterparts in nearly all cases, particularly for longer forecasting horizons (e.g., 45-min predictions). The most significant improvements are observed in RMSE, where SAEA reduces errors by 5%–10% across different forecasting models, showcasing its ability to mitigate compounding prediction errors. The empirical findings in Figs. 4 and 5 further validate the impact of SAEA in reducing error autocorrelation. Moreover, the structural coefficient analysis in Fig. 6 highlights how SAEA leverages spatial dependencies by capturing meaningful relationships between nodes, ensuring a more interpretable and data-driven error correction mechanism. These results confirm that by incorporating spatiotemporal correlation into the training objective and leveraging structured sparsity, SAEA effectively enhances both predictive accuracy and robustness, making it a promising enhancement for state-of-the-art traffic forecasting models.

## 5. Conclusion

Traffic forecasting plays a critical role in intelligent transportation systems (ITS), where deep neural network (DNN)-based models, particularly graph neural networks (GNNs), have demonstrated remarkable success in capturing complex spatiotemporal dependencies. However, existing forecasting models predominantly assume that prediction errors are independent and identically distributed (i.i.d.), which contradicts the inherent autocorrelation observed in real-world traffic data. This oversight limits the accuracy and robustness of existing models.

To address this challenge, we propose Spatiotemporally Autocorrelated Error Adjustment (SAEA), a novel framework that explicitly accounts for spatiotemporal error correlations in traffic forecasting. By modeling prediction errors using a vector autoregressive (VAR) process, we incorporate a coefficient matrix into the cost function, allowing models to jointly learn forecasting parameters and error dependencies. Additionally, structured sparsity regularization is introduced to embed prior spatial information into the coefficient matrix, ensuring that learned correlations align with the underlying road network topology.

Through extensive experiments on multiple state-of-the-art forecasting models and real-world traffic datasets, we demonstrate that SAEA significantly improves prediction accuracy while enhancing the interpretability of error correlations. Specifically, SAEA achieves an average prediction accuracy improvement of 6.5% across five models, highlighting its effectiveness in mitigating autocorrelated prediction errors. The results confirm that SAEA is a model-agnostic approach, applicable to a wide range of forecasting architectures without requiring modifications to their internal structures.

Despite its advantages, SAEA is constrained by its VAR-based linear assumption, which may not fully capture nonlinear error dependencies present in complex traffic systems. Future research could explore nonlinear extensions of the error adjustment process, leveraging deep generative models or reinforcement learning-based approaches to enhance adaptability. Furthermore, extending SAEA to multi-modal transportation networks and real-time adaptive forecasting remains an exciting direction for future exploration.

By systematically addressing spatiotemporal error correlations, SAEA provides a significant step toward more robust and interpretable traffic forecasting models, paving the way for more resilient ITS applications.

The proposed SAEA framework introduces a principled approach to adjusting autocorrelated errors, which is commonly overlooked in current deep learning-based forecasting models. In traffic forecasting, this enables more accurate and interpretable predictions by modeling how residual errors propagate across both spatial and temporal dimensions. Beyond traffic, the concept of error adjustment using structured temporal models (e.g., VAR processes) and structured sparsity can be extended to other spatiotemporal domains such as climate modeling, epidemiology, and energy consumption forecasting, where correlated residuals are also prevalent. We believe SAEA opens new avenues for incorporating statistical rigor into deep learning-based forecasting, encouraging future research to revisit common assumptions (e.g., i.i.d. errors) and to integrate classical time series insights into modern learning paradigms.

## CRediT authorship contribution statement

**Fuqiang Liu:** Writing – original draft, Software, Methodology, Data curation, Conceptualization. **Weiping Ding:** Writing – review & editing. **Luis Miranda-Moreno:** Writing – review & editing, Supervision. **Lijun Sun:** Writing – review & editing, Supervision.

## Declaration of competing interest

The authors declare that they have no known competing financial interests or personal relationships that could have appeared to influence the work reported in this paper.

## Acknowledgment

This research is supported by the Quebec Research Fund Nature and Technology under grant No. 306529 on Smart Cities and Big Data. Besides, F. Liu would like to thank FRQNT for providing the B2X Doctoral Scholarship.

## Data availability

Data will be made available on request.

**Table 3**
Effectiveness comparison of VAR(1) and VAR(2).

|  | PeMS (15/30/45 min) | | METR-LA (15/30/45 min) | |
| --- | --- | --- | --- | --- |
|  | MAPE (%) | RMSE | MAPE (%) | RMSE |
| STGCN (no adjustment) | **5.24**/7.39/9.12 | 4.06/5.74/7.01 | **7.62**/9.57/10.12 | 5.73/7.22/7.78 |
| STGCN (VAR(1)) | 5.29/**7.37**/**8.96** | **4.02**/**5.46**/**6.26** | 7.64/**9.47**/**10.07** | **5.64**/**6.92**/**7.61** |
| STGCN (VAR(2)) | 5.29/7.42/8.99 | 4.07/5.63/6.58 | 7.65/9.50/10.09 | 5.70/7.07/7.70 |
| GWave (no adjustment) | **5.21**/7.67/8.66 | 4.13/5.78/6.92 | 6.90/8.37/9.68 | 5.14/6.20/7.42 |
| GWave (VAR(1)) | 5.24/**7.61**/**8.55** | **4.11**/**5.71**/**6.24** | **6.84**/**8.33**/**9.58** | **5.12**/**6.14**/**7.15** |
| GWave (VAR(2)) | 5.25/7.64/8.58 | 4.13/5.75/6.49 | 6.90/8.36/9.65 | 5.16/6.19/7.30 |